\pdfoutput=1

\documentclass[11pt]{article}

\usepackage{acl}

\usepackage{times}
\usepackage{latexsym}

\usepackage[T1]{fontenc}

\usepackage[utf8]{inputenc}

\usepackage{microtype}

\usepackage{booktabs}
\usepackage{multirow}
\usepackage{adjustbox}
\usepackage{subcaption}
\usepackage{graphicx}
\usepackage{pifont}
\usepackage{tabularx}
\usepackage{xcolor}
\usepackage{lipsum}
\usepackage{amsmath}
\usepackage{multirow}
\usepackage[normalem]{ulem}
\usepackage{float}
\useunder{\uline}{\ul}{}
\usepackage{CJKutf8}
\usepackage{rotating}
\usepackage{nth}
\usepackage{enumitem}
\usepackage{tablefootnote}
\newcommand{\tabitem}{~~\llap{\textbullet}~~}
\usepackage{xurl}

\title{Rethinking Annotation: Can Language Learners Contribute?}

\author{
    Haneul Yoo$^1$, Rifki Afina Putri$^1$, Changyoon Lee$^1$, Youngin Lee$^1$, \\
    \textbf{So-Yeon Ahn$^1$, Dongyeop Kang$^2$, Alice Oh$^1$} \\
    $^1$KAIST, South Korea, $^2$University of Minnesota, USA \\
    \texttt{\{\href{mailto:haneul.yoo@kaist.ac.kr}{\color{black}{haneul.yoo}}, \href{mailto:rifkiaputri@kaist.ac.kr}{\color{black}{rifkiaputri}}, \href{mailto:cyoon47@kaist.ac.kr}{\color{black}{cyoon47}}, \href{mailto:conviette@kaist.ac.kr}{\color{black}{conviette}}\}@kaist.ac.kr}, \\
    \texttt{ahnsoyeon@kaist.ac.kr}, \texttt{dongyeop@umn.edu}, \texttt{alice.oh@kaist.edu}
}

\begin{document}
\maketitle
\begin{abstract}
Researchers have traditionally recruited native speakers to provide annotations for widely used benchmark datasets.
However, there are languages for which recruiting native speakers can be difficult, and it would help to find learners of those languages to annotate the data. 
In this paper, we investigate whether language learners can contribute annotations to benchmark datasets.
In a carefully controlled annotation experiment, we recruit 36 language learners, provide two types of additional resources (dictionaries and machine-translated sentences), and perform mini-tests to measure their language proficiency.
We target three languages, English, Korean, and Indonesian, and the four NLP tasks of sentiment analysis, natural language inference, named entity recognition, and machine reading comprehension.
We find that language learners, especially those with intermediate or advanced levels of language proficiency, are able to provide fairly accurate labels with the help of additional resources.
Moreover, we show that data annotation improves learners' language proficiency in terms of vocabulary and grammar. 
One implication of our findings is that broadening the annotation task to include language learners can open up the opportunity to build benchmark datasets for languages for which it is difficult to recruit native speakers.
\end{abstract}

\section{Introduction}
Data annotation is important, and in NLP, it has been customary to recruit native speakers of the target languages, even though it is difficult to recruit native speakers for many languages. Meanwhile, there are many people learning another language, for instance, Duolingo claims that 1.8 billion people are learning a foreign language using their app.\thinspace\footnote{\url{https://www.duolingo.com/}}

\begin{figure}[ht!]
    \centering
    \includegraphics[width=\columnwidth]{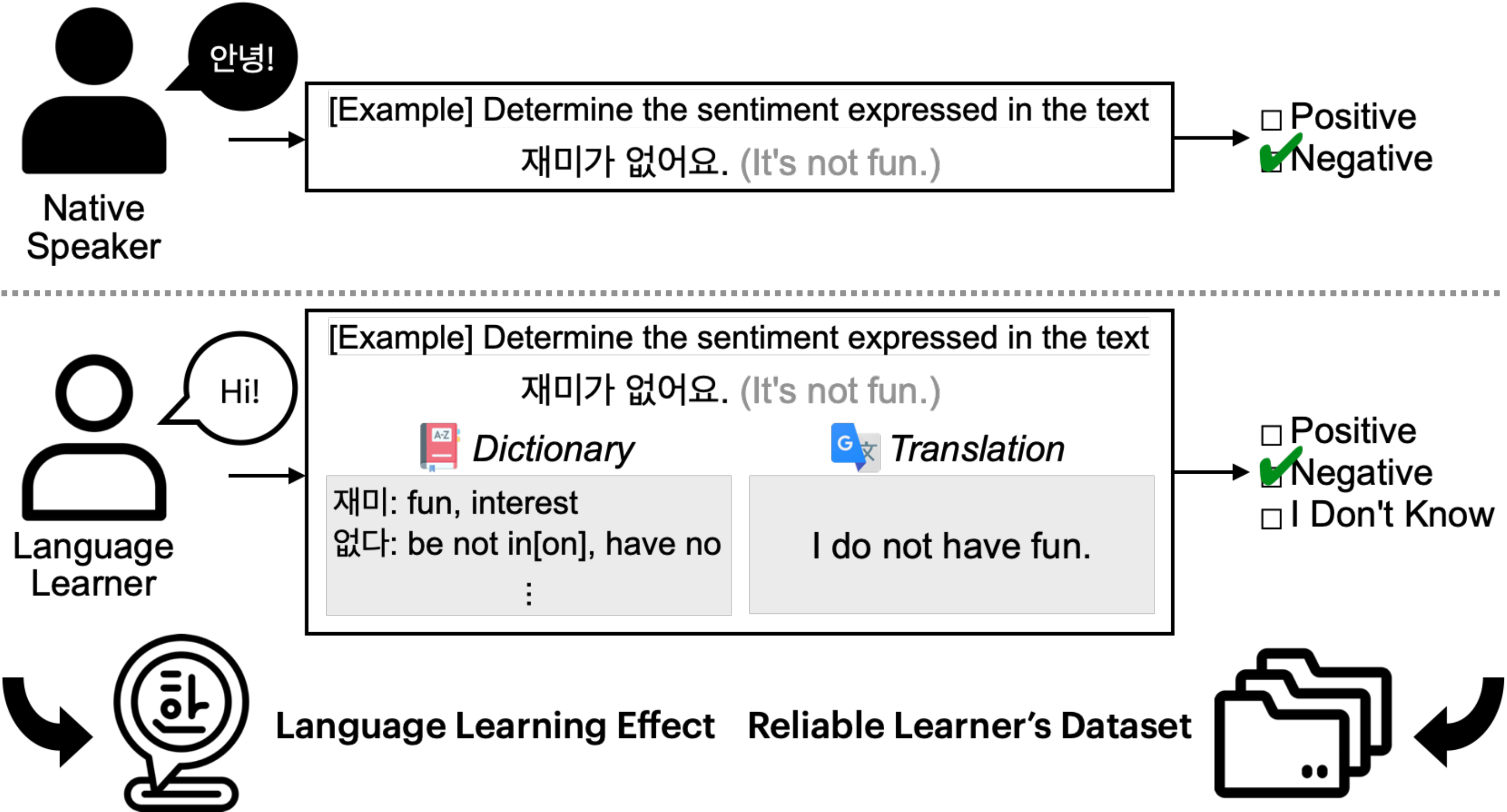}
    \caption{Recruiting language learners in NLP data annotation. They can be assisted by dictionaries or MT systems.}
    \label{fig:overview}
\end{figure}

In this paper, we examine whether language learners can annotate data as well as native speakers and whether their annotations can be used to train language models.
We explore this question with five control variables that may affect the quality of language learner annotations. These are the language, task, learners' language proficiency, difficulty of the annotation questions, and additional resources that learners can consult.
We recruited learners at various levels of proficiency in English (high-resource), Korean (mid-resource), and Indonesian (low-resource). They annotated data on four tasks, sentiment analysis (SA), natural language inference (NLI), named entity recognition (NER), and machine reading comprehension (MRC).
We ask three levels of learners to complete multiple sessions of data annotation given with the help of a dictionary or machine-translated texts.

Our major findings, both in terms of the quality and learning effect of learners' annotations, are summarized as follows:
We measure the degree of inter-annotator agreement between learners and ground truth labels, and show that \textit{language learners can annotate data at a fairly accurate level}, especially for the simpler tasks of SA and NER, and for easy- to medium-level questions.
Language learners consulting dictionaries generate more accurate labels than learners consulting machine-translated sentences.
Language models trained on data generated from the distribution of the learners' annotations achieved performance comparable to those of models trained on ground truth labels, demonstrating the efficacy of learner-annotated data.

We also observe that \textit{learners' language proficiency in vocabulary and grammar tends to improve} as they carry out the annotation tasks.
We measure their proficiency by conducting pre- and post-tests before and after the annotation.
Learners perceive that their language proficiency improved during data annotation, and most were willing to re-participate in the process.

We hope this paper allows researchers to question the necessity of recruiting native speakers for data annotation and call on other NLP researchers carefully to consider the criteria by which to recruit crowdworkers for data annotation carefully.

\section{Related Work}
We can group annotators of NLP datasets into language learners, non-speakers, and non-experts.
Language learners are people who are learning the target language, while non-speakers are those who have never learned the target language. Non-experts are people who have no expertise in NLP tasks or data annotations. We look at previous work with these three annotator groups.

\paragraph{Language Learner Annotation.}
There are several tools for both language learning and crowdsourcing that create linguistic resources.
The early motivation of Duolingo was to translate the web with language learners \citep{vonahn2013duolingo}.
\citet{hladka2014crowdsourcing} introduced a pilot experiment on Czech, the aim of which was both data annotation and the teaching of grammar.
\citet{sangati2015school} proposed a web-based platform similar to that of Duolingo that undertakes POS tagging with grammar exercises through interactions between a teacher's validation and students' annotations.
\citet{nicolas-etal-2020-creating} employed language learners to extend existing language resources (ConceptNet \citep{liu2004conceptnet}), showing that this method also has educational values.
However, they did not explicitly mention the details of their experimental settings, including the number of participants, and there was no study that recruited and employed language learners in NLP tasks with a comprehensive empirical analysis of diverse factors.

\paragraph{Non-speaker Annotation.}
A recent study employed non-speakers on specific NLP tasks and provided tools for non-speaker annotators, but that study mainly focused on easy tasks such as NER and binary classification tasks.
\citet{tsygankova-etal-2021-building} employed non-speakers as annotators to build a NER dataset and model for Indonesian, Russian, and Hindi and compared their performances with those of fluent speakers'.
The non-speakers produced meaningful results for NER in Indonesian on a combination of an easy task and an easy language written in the Latin alphabet with simple grammar.
\citet{mayhew-etal-2020-simultaneous, kreutzer2022quality} also employed non-speakers for some easy tasks such as NER along with native or fluent speakers.
Despite these efforts, it remains unclear as to whether non-speakers can undertake annotation on more complex tasks such as MRC with a paragraph to read, and NLI, requiring a comprehensive understanding of the premise and hypothesis sentences to infer the connection between the sentences correctly.

\citet{hermjakob-etal-2018-translating, mayhew-roth-2018-talen, lin-etal-2018-platforms, costello-etal-2020-dragonfly} devised assisting tools for non-speaker annotation, providing English translation, romanization, dictionary matching, and grammar-related descriptions.
We expect that English translation and dictionary matching may also be helpful to language learners and adopt the same setup.
However, neither romanization nor grammar-related descriptions may help the learners because they already have some background knowledge of the target language, unlike the non-speakers.

\paragraph{Non-expert Annotation.}
\citet{snow-etal-2008-cheap} suggested using a collection of non-expert annotations rather than expensive expert annotations.
They analyzed and compared those two types of annotations on several NLP tasks.
Only relatively few non-expert annotations are necessary to equal the performance of an expert annotator for certain simple tasks.
\citet{madge2019progression} suggest the training of non-expert annotators via progression in a language annotation game considering the linguistic ability of crowdworkers and the readability level of documents.

\section{Study Design}
This section describes how we carefully design our controlled experiments with diverse factors that may affect the quality of learners' annotations and the learning effect.

\begin{table*}[!t]
\small
\centering
\begin{tabularx}{\textwidth}{@{}ccccc@{}}
\toprule
                    & \textbf{SA}                                            & \textbf{NLI}                                           & \textbf{NER}                                                    & \textbf{MRC}                         \\ \midrule
\multirow{2}{*}{EN} & \multicolumn{1}{c}{SST2}                               & \multicolumn{1}{c}{SNLI}                               & \multicolumn{1}{c}{CoNLL++}                                             & \multicolumn{1}{c}{TyDiQA}                       \\
                    & \multicolumn{1}{c}{\citep{socher-etal-2013-recursive}} & \multicolumn{1}{c}{\citep{young-etal-2014-image}}      & \multicolumn{1}{X}{\citep{tjong-kim-sang-de-meulder-2003-introduction}} & \multicolumn{1}{c}{\citep{clark-etal-2020-tydi}} \\
\multirow{2}{*}{KO} & \multicolumn{1}{c}{NSMC}                               & \multicolumn{1}{c}{KLUE}                               & \multicolumn{1}{c}{KLUE}                                                & \multicolumn{1}{c}{TyDiQA}                       \\
                    & \multicolumn{1}{c}{\citep{park2016nsmc}}               & \multicolumn{1}{c}{\citep{park2021klue}}               & \multicolumn{1}{c}{\citep{park2021klue}}                                & \multicolumn{1}{c}{\citep{clark-etal-2020-tydi}} \\
\multirow{2}{*}{ID} & \multicolumn{1}{c}{IndoLEM}                            & \multicolumn{1}{c}{IndoNLI}                            & \multicolumn{1}{c}{NERP}                                                & \multicolumn{1}{c}{TyDiQA}                       \\
                    & \multicolumn{1}{c}{\citep{koto-etal-2020-indolem}}     & \multicolumn{1}{c}{\citep{mahendra-etal-2021-indonli}} & \multicolumn{1}{c}{\citep{wilie-etal-2020-indonlu}}                     & \multicolumn{1}{c}{\citep{clark-etal-2020-tydi}} \\ \bottomrule
\end{tabularx}
\caption{Source dataset for each language and task.}
\label{tab:source_data}
\end{table*}

\begin{table}[!t]
\centering
\resizebox{\columnwidth}{!}{
\begin{tabular}{@{}lr@{}}
\toprule
\textbf{Control Variables} & \textbf{Values}                \\ \midrule
Language                   & EN, KO, ID                     \\
Task                       & SA, NLI, NER, MRC              \\
Learner Fluency            & Basic, Intermediate, Advanced  \\
Question Difficulty        & Very easy, $\cdots$, Very hard \\
Additional Resources       & Dictionary, Translation        \\ \bottomrule
\end{tabular}
}
\caption{Control variables in our experiments.}
\label{tab:control_variables}
\end{table}

\subsection{Control Variables}

Table \ref{tab:control_variables} shows a summary of the different control variables considered in our experiments with the corresponding values.
We should take these control variables into account when simulating learners' annotations in real-world scenarios and use diverse combinations of them.
We set the major control variables based on previous work on NLP data annotation \citep{joshi-etal-2020-state, wang-etal-2018-glue, lin-etal-2018-platforms} and language learning \citep{lee2006from, crossley2008assessing, shieh2010using}.

\paragraph{Language Selection.}
We choose three target languages, English (EN), Korean (KO), and Indonesian (ID), based on the availability of gold-label data, the availability of native speakers to evaluate, and the difficulty of the language.
English is the highest-resource language, while Korean and Indonesian are mid- to low-resource languages, respectively \citep{joshi-etal-2020-state}.
Korean uses its own alphabet, while Indonesian adopts the Latin alphabet.
The Foreign Service Institute (FSI)\thinspace\footnote{\url{https://www.state.gov/foreign-language-training/}} categorizes languages into five categories based on the amount of time it takes to learn them considering several variables, including grammar, vocabulary, pronunciation, writing system, idiomatic expressions, distance from English, dialects, and learning resources. According to the FSI ranking, Indonesian is in category 2, requiring around 36 weeks or 900 class hours, Korean is in category 4, requiring 88 weeks or 2200 class hours to reach B2/C1 level in CEFR, and English is in category 0.

\paragraph{Task and Data.}

We choose four tasks from each common task type in the GLUE benchmark \citep{wang-etal-2018-glue}: sentiment analysis (SA) for single sentence classification, natural language inference (NLI) for sentence pair classification, named entity recognition (NER) for sequence tagging, and machine reading comprehension (MRC) for span prediction.
Table \ref{tab:source_data} presents a list of the datasets used in our study.
SA has two options (\texttt{positive} and \texttt{negative}), and NLI has three options (\texttt{entailment}, \texttt{neutral}, and \texttt{contradict}) for all languages.
The NER datasets have different categories of named entities among the languages, while all languages have \texttt{person} and \texttt{location} entities.

\paragraph{Participant Selection.}

We adopt and revise the CEFR\thinspace\footnote{Common European Framework of Reference for Languages (\url{https://www.coe.int/en/web/common-european-framework-reference-languages})} criteria to categorize learners into three levels: basic (A1-A2), intermediate (B1-B2), and advanced (C1-C2).
Table \ref{tab:learner_level_criteria} shows our recruiting criteria with respect to language fluency.
We do not request official test scores for basic-level learners, as they may not have taken official language proficiency tests.
We assign the learners at each level to annotate questions to facilitate majority voting among three responses from different levels of participants.
All annotators in our experiments are non-experts in NLP data annotations, and three annotators are allocated to each task and each additional resource.
Participants are asked to do two tasks: SA and MRC, or NER and NLI.
The study involved participants with ages ranging from 19 to 44 (average 31.5, median 24) at the time of the experiment. They are primarily undergraduate or graduate students, with some office workers and unemployed individuals.

\begin{table*}[!t]
\centering
\resizebox{\linewidth}{!}{%
\begin{tabular}{@{}llll@{}}
\toprule
\multicolumn{1}{c}{\textbf{}} & \multicolumn{1}{c}{\textbf{Basic}}              & \multicolumn{1}{c}{\textbf{Intermediate}}        & \multicolumn{1}{c}{\textbf{Advanced}} \\ \midrule
EN                            & Self report A                               & \texttt{TOEFL}\thinspace\footnotemark 57-109                          & \texttt{TOEFL} $\geq$ 110               \\
KO                            & Learning experience < 1 yr \& Self report A & \texttt{TOPIK} Level 2-4                       & \texttt{TOPIK} $\geq$ Level 5           \\
ID                            & Learning experience < 1 yr \& Self report A & \texttt{OPI}\thinspace\footnotemark $\leq$ IH || \texttt{FLEX}\thinspace\footnotemark $\approx$ 600   & \texttt{OPI} $\geq$ AL || \texttt{TIBA}\thinspace\footnotemark $\geq$ 4 \\
\bottomrule
\end{tabular}%
}
\caption{Learner level criteria}
\label{tab:learner_level_criteria}
\end{table*}

\paragraph{Additional Resources.}
\citet{lin-etal-2018-platforms} observed that additional resources such as dictionary matching or English translation may assist non-speakers with annotation tasks.
We divide the participants into two groups with the additional resources at their disposal, in this case a dictionary and translations provided by a commercial MT system.
We only provide texts in the target language and ask participants to consult online or offline dictionaries if they need any help in the dictionary setting.
Otherwise, we provide both the texts in the target language and corresponding translations created by the Google Translate API on our website and ask the participants not to use any other external resources.

\paragraph{Annotation Sample Selection.}
We randomly sample 120 annotation samples for each task from the source datasets and categorize them into five groups based on their difficulty level.
The sentence-level difficulty score is calculated using a macro average of several linguistic features from Coh-Metrix \cite{graesser2004coh}, a metric for calculating the coherence and cohesion of texts. The linguistic features that we use in our experiment are the \textit{lexical diversity}, \textit{syntactic complexity}, and \textit{descriptive measure}.
Lexical diversity is computed by the type-token ratio, syntactic complexity is computed according to the number of conjunction words, and descriptive measure is computed by the sentence character length, the number of words, and the mean of the number of word syllables.
We add additional metrics for MRC tasks that contain a paragraph, in this case the number of sentences in the paragraph, the character length of the answer span, and the number of unique answers. The paragraph-level difficulty score is calculated by taking the average of the sentence-level scores in the paragraph.

\paragraph{Test Question Selection.}
Pre- and post-tests are used, consisting of five questions from official language proficiency tests and ten questions asking about the meanings of words appearing in annotation samples that they will solve in the same session.
Standardized test questions explore whether participating in the annotation improves the learners' overall language proficiency over several days, while word meaning questions aim to inspect whether participating in the annotation helps them learn some vocabulary.

\addtocounter{footnote}{-4}
\addtocounter{footnote}{1}
\footnotetext{Test Of English as a Foreign Language (\url{https://www.ets.org/toefl})}
\addtocounter{footnote}{1}
\footnotetext{Oral Proficiency Interview (\url{https://www.actfl.org/assessment-research-and-development/actfl-assessments/actfl-postsecondary-assessments/oral-proficiency-interview-opi})}
\addtocounter{footnote}{1}
\footnotetext{Foreign Language EXamination (\url{https://www.kotga.or.kr/sub/sub03_01.php})}
\addtocounter{footnote}{1}
\footnotetext{Tes Bahasa Indonesia sebagai Bahasa Asing (\url{https://lbifib.ui.ac.id/archives/105})}

We use \texttt{TOPIK}\thinspace\footnote{Test Of Proficiency In Korean (\url{https://www.topik.go.kr/})} for Korean, \texttt{UKBI}\thinspace\footnote{Uji Kemahiran Berbahasa Indonesia (\url{https://ukbi.kemdikbud.go.id/})} and \texttt{BIPA}\thinspace\footnote{Bahasa Indonesia untuk Penutur Asing (\url{https://bipa.ut.ac.id/})} for Indonesian, and \texttt{TOEIC}\thinspace\footnote{Test Of English for International Communication (\url{https://www.ets.org/toeic})} and \texttt{GRE}\thinspace\footnote{Graduate Record Examination (\url{https://www.ets.org/gre})} for English.
We chose nouns and verbs from annotation questions and created multiple-choice questions whose answers are the nouns or the verbs in the annotation questions.

\begin{table*}[t!]
\centering
\begin{tabular}{@{}llccc@{}}
\toprule
                                   &             & \textbf{Accuracy} & \textbf{Inter-Annotator Agreement} & \textbf{Time} (min)\\ \midrule
Native Speakers                    & -           & 8.53$_{\pm 0.09}$ & 0.77$_{\pm 0.02}$                  & 4.07$_{\pm 0.78}$   \\
\cline{1-2}\cline{3-5}
\multirow{2}{*}{Language Learners} & Dictionary  & 7.72$_{\pm 0.09}$ & 0.70$_{\pm 0.01}$                  & 6.92$_{\pm 0.70}$   \\
                                   & Translation & 7.31$_{\pm 0.09}$ & 0.67$_{\pm0.01}$                   & 6.49$_{\pm 0.36}$   \\ \bottomrule
\end{tabular}
\caption{
Annotation comparison between native speakers and learners (with dictionary and translation settings).
Accuracy means the number of correct questions compared to the ground truth labels out of 10. Inter-Annotator agreement means pairwise F1-score. Time means how long annotating 10 samples takes in minutes.}
\label{tab:general_results}
\end{table*}

\subsection{Workflow}

\begin{figure}[t!]
    \centering
    \includegraphics[width=.8\columnwidth]{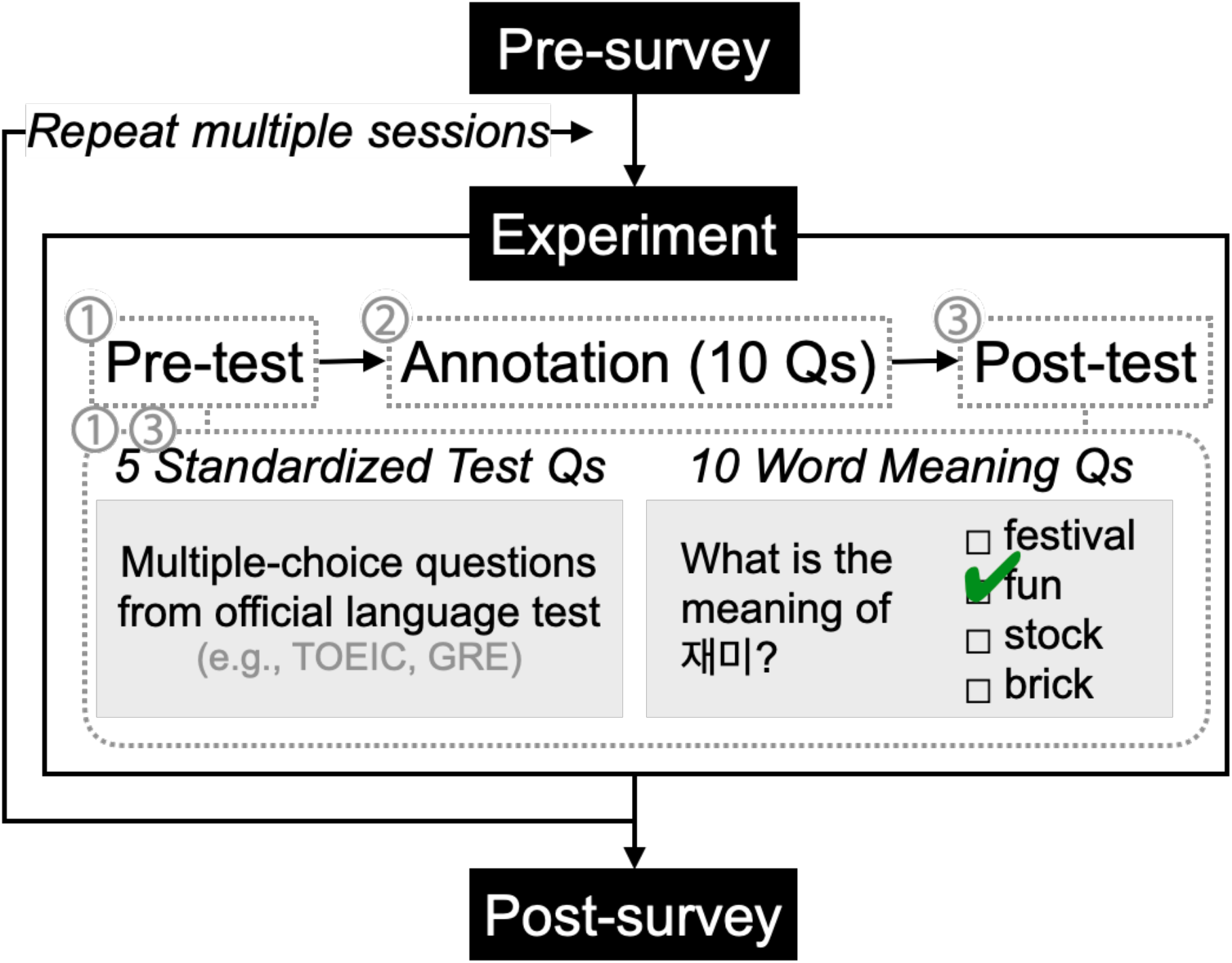}
    \caption{High-level flowchart of our experiments.}
    \label{fig:flowchart}
\end{figure}

\paragraph{Step 1: Pre-survey}
As shown in Figure \ref{fig:flowchart}, we use a survey to ask participants about their self-rated language fluency, language background, and learning experience before the main experiments.
We describe the CEFR criteria and ask participants to self-evaluate their language proficiency in general, colloquial, and formal texts and choose which of the colloquial and formal texts they are more familiar with.

\paragraph{Step 2: Experiment}
Our experiments consist of a series of multiple sessions over six days.
Each session consists of three steps, and we ask participants to do two sessions per task per day and repeat this for six consecutive days.
Before starting the main experiment, we provide a pilot session to check whether the participants fully understand our instructions.
All of the experimental processes are done on our research website, and we measure the time spent by the participants on each step.

\paragraph{Step 2.1: Pre-test}
Participants solve 15 test questions to check their language proficiency level.
All test questions are multiple-choice types and include the ``\textit{I don't know}'' option.

\paragraph{Step 2.2: Annotation}
Participants annotate ten questions with the help of the additional resources assigned.

\paragraph{Step 2.3: Post-test}
After completing the annotation, participants solve the same 15 test questions they solved in the pre-test.
This step investigates whether data annotation has any learning effect.

\paragraph{Step 3: Post-survey}
After the experiments, participants complete a post-survey about their thoughts on annotation and self-rated language proficiency.
They answer the questions below for each task on a five-point Likert scale from ``\textit{strongly disagree}'' to ``\textit{strongly agree}''.



\section{Experimental Results}

We discuss the results of our experiments with respect to two research questions:
\begin{enumerate}
    \item Can we obtain a reliable dataset from learners' annotations? Which design setting would be most helpful?
    We answer this question via quality assessment (\S\ref{sec:results_annotation_quality}), training simulation (\S\ref{sec:results_simulation}), and error analysis (\S\ref{sample-level_analysis}).
    \item Do learners improve their language proficiency while annotating the NLP tasks (\S\ref{sec:discussion_learning_effect})? 
    
\end{enumerate}

All findings we discuss in this section were shown to be statistically significant at $p$ level of  $< 0.05$ using ANOVA. Specifically, comparisons for annotation accuracy, annotation time, and survey responses were analyzed with four-way ANOVA over the four between-subject factors of task, language, additional resources, and learner level. Comparisons between pre-test and post-test results were done with a mixed two-way ANOVA with learner level and additional resources as between-subject factors. Pairwise t-tests were conducted for all factors with Bonferroni corrections.

\subsection{Annotation Quality}\label{sec:results_annotation_quality}

\paragraph{Accuracy and Agreement.}

Table \ref{tab:general_results} shows the results of annotations generated by language learners compared to native speakers.
Language learners made correct annotations to 7.48 questions among 10 questions on average\thinspace\footnote{Annotation accuracy was computed by a weighted averaged F1 score compared to the ground truth label on NER and MRC. The average of the weighted-averaged F1 score was used for some samples in MRC with multi-choice answers.}, taking 6.68 minutes.
They generated 1.05 less accurate labels and took 2.6 minutes longer time than the native speakers.
Learners assisted by dictionaries can produce more reliable labels than learners using MT system.
Meanwhile, majority voting among native speakers generated 19 incorrect labels out of 120 questions, compared to learners’ 21.5 incorrect labels (Table \ref{tab:aggregation_majority_vote} in Appendix).
This shows that language learners’ annotations can be aggregated by majority voting to be nearly as accurate as those of native speakers.

\paragraph{Languages and Tasks.}

\begin{figure}[!ht]
    \centering
    \subfloat{\includegraphics[width=\columnwidth]{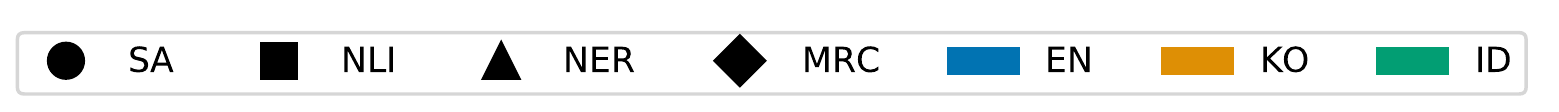}}
    \addtocounter{subfigure}{-1}
    \subfloat[\centering Annotation accuracy]{\includegraphics[height=4.5cm]{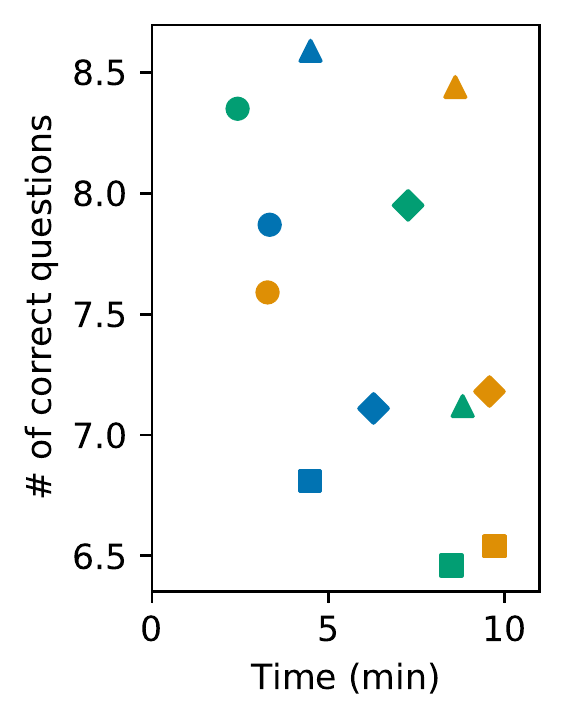}}
    \subfloat[\centering Inter-annotator agreement]{\includegraphics[height=4.5cm]{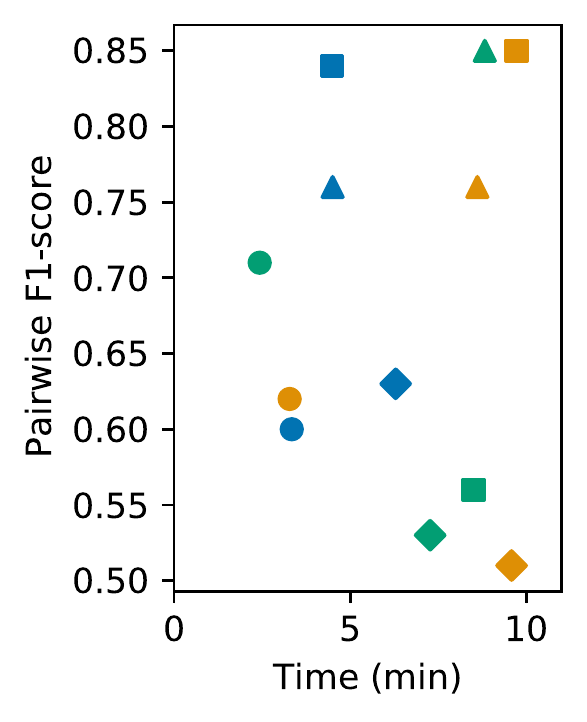}}
    \caption{Task difficulty according to language and task.}
    \label{fig:task_difficulty}
\end{figure}

Figure \ref{fig:task_difficulty} (a) and (b) show the task difficulty with respect to time versus annotation accuracy and inter-annotator agreement, respectively.
SA and NER are easier for language learners than NLI and MRC, considering both accuracy and time.
MRC, which requires paragraph comprehension, unlike sentence-level tasks, may be difficult for learners.
Nonetheless, they achieved high accuracy, and most of their answer spans overlapped with the ground truth answers.
Detailed results and further analysis of the outcomes in Figure \ref{fig:task_difficulty} can be found in Appendix \ref{appendix:annotation_accuracy}.

We measure inter-annotator agreement using the pairwise F1 scores.
Table \ref{tab:task_difficulty} (b) shows the level of agreement and the standard error for each language and task.
Both NLI and NER show high agreement, while the token-based task MRC shows relatively low agreement compared to the other tasks.

Korean SA shows low agreement, most likely due to some noisy samples in the NSMC dataset.
The NSMC dataset is a movie review dataset whose negative labels come from the reviews with ratings of 1-4, and where the positive labels come from those with ratings of 9-10, respectively.
This dataset contains noisy samples whose gold labels are unreliable or whose labels cannot be determined only with the text, requiring some metadata.

MRC in Korean shows low agreement, and we assume this stems from the fact that Korean is a morpheme-based language while the others use word-based tokenization.
The F1 score was computed based on the corresponding word overlaps in both English and Indonesian. Korean uses character-based overlap, which is stricter.
It may be more complicated for annotators to clearly distinguish the answer span at the character level rather than at the word level.

\paragraph{Language Proficiency and Question Difficulty.}

\begin{figure}[t!]
    \centering
    \includegraphics[width=0.9\columnwidth]{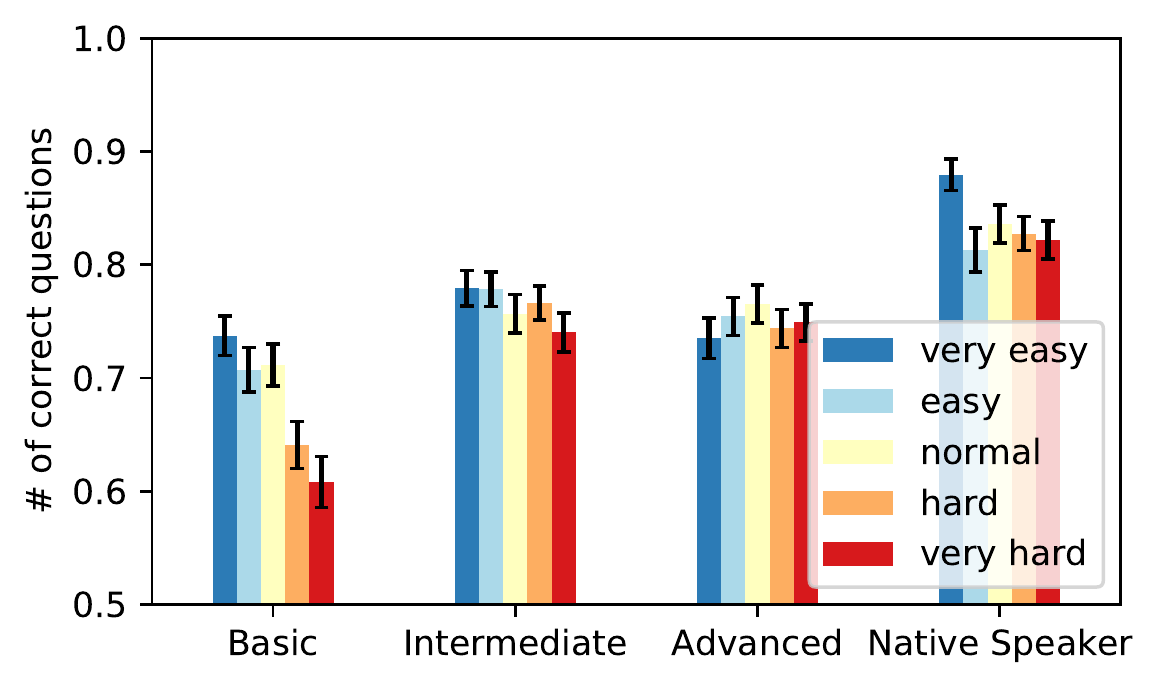}
    \caption{Annotation accuracy on question difficulty and language fluency.}
    \label{fig:question_level_vs_learner_level}
\end{figure}

\begin{table*}[t!]
\centering
\resizebox{\textwidth}{!}{
\begin{tabular}{@{}llcccccc@{}}
\toprule
                                   &             & \multicolumn{3}{c}{\textbf{SA}}                              & \multicolumn{3}{c}{\textbf{NLI}}                             \\ \cmidrule(l){3-5} \cmidrule(l){6-8} 
                                   &             & EN                 & KO                 & ID                 & EN                 & KO                 & ID                 \\ \midrule
Ground Truth                       & -           & 89.56$_{\pm 1.11}$ & 85.29$_{\pm 0.79}$ & 97.20$_{\pm 0.86}$ & 79.05$_{\pm 1.44}$ & 79.00$_{\pm 2.48}$ & 68.20$_{\pm 1.32}$ \\
MT Dataset                         & -           & 79.25$_{\pm 1.25}$ & 75.27$_{\pm 1.33}$ & 87.19$_{\pm 1.39}$ & 56.78$_{\pm 2.26}$ & 47.06$_{\pm 1.26}$ & 52.35$_{\pm 1.35}$ \\ \midrule
Native Speakers                    & -           & 87.59$_{\pm 1.30}$ & 89.18$_{\pm 1.62}$ & 94.18$_{\pm 0.31}$ & 71.86$_{\pm 1.43}$ & 74.09$_{\pm 1.41}$ & 67.21$_{\pm 1.23}$ \\ \midrule
\multirow{3}{*}{Language Learners} & All         & 89.09$_{\pm 1.87}$ & 89.26$_{\pm 1.41}$ & 94.26$_{\pm 0.73}$ & 72.16$_{\pm 1.91}$ & 71.82$_{\pm 1.27}$ & 70.39$_{\pm 2.17}$ \\
                                   & Dictionary  & 86.64$_{\pm 0.40}$ & 87.61$_{\pm 1.06}$ & 92.61$_{\pm 1.44}$ & 70.40$_{\pm 1.09}$ & 74.22$_{\pm 1.98}$ & 66.70$_{\pm 1.48}$ \\
                                   & Translation & 85.39$_{\pm 1.14}$ & 87.47$_{\pm 1.65}$ & 92.47$_{\pm 1.46}$ & 74.69$_{\pm 1.63}$ & 73.03$_{\pm 1.02}$ & 69.84$_{\pm 2.49}$ \\ \bottomrule
\end{tabular}
}
\caption{
Training simulation of annotations by native speakers and learners.
BERT-based models are trained on labels generated or synthesized by each group.
We provide the upper and lower bounds on the performances based on ground-truth labels and translations, respectively. 
}
\label{tab:synthetic_result_majority_vote}
\end{table*}

Figure \ref{fig:question_level_vs_learner_level} shows the percentage and the standard error of obtaining a correct answer for each question difficulty and learner fluency.
Both intermediate and advanced learners show similar levels of accuracy regardless of question difficulty level, while basic-level learners tend to fail on complex questions.
The mean number of correct questions out of 10 increases to 7.66 without basic-level learners.
This implies that the intermediate level is sufficient to understand the sentences in general NLP datasets and suggests the feasibility of recruiting learners as annotators in place of native speakers, especially on easy-to-medium tasks and questions.

\subsection{Training Simulation with Learners' Annotations}\label{sec:results_simulation}

In order to show the reliability of learners' annotations used as training labels for language models, we compare the performance of models trained on learners' annotations across SA and NLI to the models trained on native speakers' annotations.
Because we only have a small number of learners' annotations, we generate synthetic data following the distribution of learners' annotations.
We randomly select 10K samples from the training data of the original datasets and change the labels into the generated synthetic labels.
We aggregate learners' annotations using a majority vote.
We ran the Shapiro-Wilk test and found that the distribution of labels is Gaussian (p-value $<$ 0.05).
We then fit the probability distributions of labels for each class and generate synthetic labels for existing NLP datasets based on those distributions.
The same process is used to build synthetic data representing native speakers' annotations.
We set two baselines as the upper and lower bounds of LMs: models trained on the original ground truth labels (Ground Truth) and models trained on machine-translated texts of other languages (MT Dataset).

We fine-tuned BERT\textsubscript{BASE} \citep{devlin-etal-2019-bert}, KLUE-BERT\textsubscript{BASE} \citep{park2021klue}, and IndoBERT\textsubscript{BASE} \citep{wilie-etal-2020-indonlu} for English, Korean, and Indonesian, respectively.
Table \ref{tab:synthetic_result_majority_vote} shows the experimental results of the LMs trained on different synthetic labels, averaged for each language.
Ground Truth indicates LMs trained on the original label, which was annotated by native speakers and merged into one by majority vote.
Models trained on synthetic labels representing learners' annotations significantly outperformed the MT Dataset.
This implies that building datasets with learners' annotation can produce more reliable labels than the baseline method of using machine-translated high-resource language datasets.

\section{Discussion}
\subsection{Qualitative Analysis on Learners' Annotation}
\label{sample-level_analysis}

\begin{table}[]
\small
\centering
\begin{tabular}{@{}ll@{}}
\toprule
\textbf{Task} & \textbf{Top-3 Failure Reasons}                                                                                                                                                \\ \midrule
SA            & \begin{tabular}[c]{@{}l@{}}\tabitem{Unreliable gold label}\\ \tabitem{Lack of background information}\\ \tabitem{Ungrammatical sentence\thinspace\footnotemark}\end{tabular}                \\ \midrule
NLI           & \begin{tabular}[c]{@{}l@{}}\tabitem{Task ambiguity}\\ \tabitem{Unreliable gold label}\\ \tabitem{Domain-specific genre and expression}\end{tabular}                           \\ \midrule
MRC           & \begin{tabular}[c]{@{}l@{}}\tabitem{Culturally-nuanced expression}\\ \tabitem{Ambiguous questions with multiple answers}\\ \tabitem{Low overlaps in answer span}\end{tabular} \\ \bottomrule
\end{tabular}
\caption{Main failure reasons on each task}
\label{tab:failure_reason}
\end{table}

\footnotetext{e.g., missing period, missing spacing and blank, nominalization, and use of slang}

We analyze the annotation result of each sample, especially the samples on which learners or native speakers failed, i.e., those that were incorrectly labeled or for which ``\textit{I don't know}'' was selected as the answer.
Table \ref{tab:failure_reason} shows the main failure reasons why learners failed to make correct annotations on each task for the samples that at most one learner correctly labeled.
The number of samples for which all learners failed ranges from zero to three for all tasks, except for NER, where no sample was incorrectly predicted by all learners; i.e., there was at least one learner who answered correctly for each question for all 120 samples. 

We found that the incorrectly labeled samples in SA mostly occurred due to the unreliable gold label in the dataset. With regard to NLI, all incorrect samples resulted from ambiguities in the task itself.
Some NLI and MRC samples are tricky for learners in that they can create correct labels only when they fully understand both the hypothesis and the premise or both the context and the question.
Fluency in English may affect failures by Indonesian learners in the translation setting, considering that the provided translations were in English.
\textit{Very difficult} examples in MRC occasionally include difficult and culturally-nuanced phrases and require background knowledge, which can be difficult for learners.

A detailed explanation of the failure reason analyses results is provided in Table \ref{tab:sample_level_analysis_example} in the Appendix.
For instance, a missing period between two short sentences, \textit{\begin{CJK}{UTF8}{mj}스토리가 어려움\end{CJK} (The story is difficult)} and \textit{\begin{CJK}{UTF8}{mj}볼만함\end{CJK} ([but it's] worth watching.)}, in Table \ref{tab:sample_level_analysis_example} (a) leads to misunderstandings among learners.
Also, an ambiguity of NLI whether ``\textit{people}'' and ``\textit{some people}'' in premise (\textit{People standing at street corner in France.}) and hypothesis (\textit{Some people are taking a tour of the factory.}) are indicating the same leads all learners and native speakers to get confused between \texttt{neutral} and \texttt{contradiction}, which is an ambiguity of NLI itself (Table \ref{tab:sample_level_analysis_example} (b)).

\subsection{Learning Effect}
\label{sec:discussion_learning_effect}

\begin{table}[t!]
\centering
\begin{subtable}{\linewidth}\centering
    \resizebox{\columnwidth}{!}{
    \begin{tabular}{@{}lccc@{}}
    \toprule
              & \textbf{Basic}        & \textbf{Intermediate}        & \textbf{Advanced}        \\ \midrule
    pre-test  & 2.72$_{\pm 0.09}$ & 3.68$_{\pm 0.08}$ & 3.99$_{\pm 0.07}$ \\
    post-test & 2.76$_{\pm 0.09}$ & 3.62$_{\pm 0.08}$ & 4.01$_{\pm 0.06}$ \\ \bottomrule
    \end{tabular}
    }
    \caption{Number of correct standardized test questions out of 5}
\end{subtable}%

\vspace{2mm}

\begin{subtable}{\linewidth}\centering
    \resizebox{\columnwidth}{!}{
    \begin{tabular}{@{}lccc@{}}
    \toprule
              & \textbf{Basic}        & \textbf{Intermediate}        & \textbf{Advanced}        \\ \midrule
    pre-test  & 7.29$_{\pm 0.12}$ & 8.93$_{\pm 0.08}$ & 9.32$_{\pm 0.06}$ \\
    post-test & 8.41$_{\pm 0.11}$ & 9.27$_{\pm 0.07}$ & 9.42$_{\pm 0.06}$ \\ \bottomrule
    \end{tabular}
    }
    \caption{Number of correct word meaning questions out of 10}
\end{subtable}
\caption{Pre-/post-test score in the same session}
\label{tab:word_meaning_questions_pre_post_test_b}
\end{table}

\paragraph{Standardized Test Questions.}
We compared pre- and post-test scores for the standardized questions in Table \ref{tab:word_meaning_questions_pre_post_test_b} (a).
There was no significant difference, implying that annotating several questions had little impact on learning grammar, structure, or general language skills in the short term.

\paragraph{Word Meaning Questions.}
Table \ref{tab:word_meaning_questions_pre_post_test_b} (b) shows the scores of the pre-/post-tests on the word meaning questions out of 10 questions.
The learning effect on vocabulary was maximized with beginner-level learners.
Both intermediate and advanced learners achieved a mean score of about 9 out of 10 on the pre-test, implying that words used in the data annotation sentences were accessible and understandable enough for them.

\paragraph{Long-term Learning Effect.}
The pre-test score for the last session is higher than that for the first session by about 4\% and 7\% each on both standardized test questions and word meaning questions, respectively (Table \ref{tab:day-by-day_pre-test}).
The increase in the standardized test question scores implies learners' improvement on general language proficiency factors, including structure and grammar.
Also, we can surmise that the vocabulary or expressions used in the NLP datasets are primarily redundant and repetitive, considering that only a few sessions can lead to an increase in pre-test scores.

\begin{table}[t!]
\centering
\begin{subtable}{\linewidth}\centering
    \begin{tabular}{@{}lccc@{}}
    \toprule
            & \textbf{Basic}        & \textbf{Intermediate}        & \textbf{Advanced}        \\ \midrule
    \nth{1} & 3.23$_{\pm 0.02}$ & 3.26$_{\pm 0.02}$ & 3.30$_{\pm 0.02}$ \\
    last    & 3.43$_{\pm 0.03}$ & 3.46$_{\pm 0.03}$ & 3.53$_{\pm 0.02}$ \\ \bottomrule
    \end{tabular}
    \caption{Number of correct standardized test questions out of 5}
\end{subtable}%

\vspace{2mm}

\begin{subtable}{\linewidth}\centering
    \begin{tabular}{@{}lccc@{}}
    \toprule
            & \textbf{Basic}        & \textbf{Intermediate}        & \textbf{Advanced}        \\ \midrule
    \nth{1} & 8.20$_{\pm 0.03}$ & 8.23$_{\pm 0.03}$ & 8.30$_{\pm 0.03}$ \\
    last    & 8.91$_{\pm 0.03}$ & 8.95$_{\pm 0.03}$ & 9.00$_{\pm 0.03}$ \\ \bottomrule
    \end{tabular}
    \caption{Number of correct word meaning questions out of 10}
\end{subtable}
\caption{Pre-test score of the first and the last session}
\label{tab:day-by-day_pre-test}
\end{table}

\subsection{Concerns about Learners' Annotation in Low-resource Languages}

This paper suggests recruiting language learners as crowdworkers in data annotation in low-resourced languages by proving the quality of learners' labels.
There are clearly many low-resource languages for which the absolute number of native speakers is exceptionally small compared to learners or for which it is almost impossible to find native speakers in the locations where NLP research is active.
For instance, we can think of endangered languages such as Irish, which has no monolingual native speaker and extremely few daily-using L1 speakers (73K) but more than 1M learners.
We can also count local languages, such as Sundanese in Indonesia and Jejueo in Korea, that are spoken by the elderly in the community, with the younger speakers who are not fluent but who are much more accessible to the researchers for annotation.

We may use either MT systems such as Google Translate considering that it supports 133 languages including several low-resource languages\thinspace\footnote{\url{https://github.com/RichardLitt/low-resource-languages}} or dictionaries for extremely low-resource languages such as Ojibwe People's Dictionary\thinspace\footnote{\url{https://ojibwe.lib.umn.edu/}}.
For low-resource languages, it is necessary to scrape together whatever resources are accessible, regardless of whether these are (incomplete) dictionaries, semi-fluent speakers, and/or anyone willing to learn and annotate in that language.

\section{Conclusion}

\begin{table*}[th!]
\centering
\begin{tabular}{@{}lccc@{}}
\toprule
                                   &             & \textbf{Time / Session (min)}    & \textbf{Expected Hourly Wage} \\ \midrule
Native Speakers                    & -           & 8.08$_{\pm 0.89}$                & KRW 9,282                     \\ \midrule
\multirow{2}{*}{Language Learners} & Dictionary  & 14.76$_{\pm 1.29}$               & KRW 20,325                    \\
                                   & Translation & 13.20$_{\pm 0.73}$               & KRW 22,727                    \\ \bottomrule
\end{tabular}
\caption{Expected hourly wage of each experiment. All wages are over the minimum wage in the Republic of Korea (KRW 9,160).}
\label{tab:expected_hourly_wage}
\end{table*}

This study provides interesting results both for the actual dataset annotation as well as understanding the non-native speakers’ annotation capabilities.
We show (1) labels provided by language learners are nearly as accurate, especially for easier tasks, (2) with additional experiments of aggregating their labels, learners’ are almost on par with native speakers, and (3) language models trained on learners’ less accurate labels achieved 94.44\% of ground truth performance.

By showing that NLP annotation does not require finding native speakers, we show the possibility of broadening NLP research for more languages, as it is very challenging to recruit native speakers for many languages.
Requiring native speakers for annotation can mean traveling to remote locations and working with an older, less-technology-savvy population.
We show that it is possible to work with language learners to hurdle geographic and technological barriers when attempting to build annotated NLP datasets.
We believe learners with high motivations and learning effects are more likely to be engaged in data annotation.

\section*{Limitations}

This paper covers only four NLP tasks.
Certain other tasks requiring more background knowledge may show different results.
We suggest recruiting language learners when native speakers are not available, but recruiting learners may also be difficult for languages that are not popular for learners.
Our results are based on a relatively low number of participants, as we chose to cover three different languages to show generalizability across languages.
Many factors that may contribute to the results remain, such as the order of the batch of annotation questions with respect to the question difficulty level.

\section*{Ethics Statement}

All studies in this research project were performed under KAIST Institutional Review Board (IRB) approval.
We consider ethical issues in our experiments with language learners and native speakers.

The first consideration is fair wages.
We estimated the average time per session (Step 2.1 to 2.3) based on a small pilot study and set the wage per session to be above the minimum wage in the Republic of Korea (KRW 9,160 $\approx$ USD 7.04)\thinspace\footnote{\url{https://www.minimumwage.go.kr/}}.
Table \ref{tab:expected_hourly_wage} shows that the expected hourly wages of all experiments exceed the minimum wage.
We estimated the time for watching the orientation video and reading the instruction manual as one hour and provided compensation for this time of KRW 10,000.

There was no discrimination when recruiting and selecting the participants for the experiment, including all minority groups and factors such as age, ethnicity, disability, and gender.
We used the sentences from publicly available datasets and manually excluded samples that may contain toxic and/or controversial contents.

\section*{Acknowledgements}
This work was supported by Institute of Information communications Technology Planning Evaluation (IITP) grant funded by the Korea government(MSIT) (No. 2022-0-00184, Development and Study of AI Technologies to Inexpensively Conform to Evolving Policy on Ethics).
This work was supported by a grant of the KAIST-KT joint research project through AI2XL Laboratory, Institute of convergence Technology, funded by KT [G01220613, Investigating the
completion of tasks and enhancing UX]. 
Rifki Afina Putri was supported by Hyundai Motor Chung Mong-Koo Global Scholarship.

\bibliography{anthology,custom}
\bibliographystyle{acl_natbib}

\clearpage

\appendix
\section*{Appendix}
\section{Experiment Setup}
\subsection{Workflow}
\paragraph{Post-survey Questions}
\begin{itemize}[leftmargin=*,noitemsep,topsep=0pt]
    \item This task is difficult for me.
    \item I think my vocabulary skills have improved after doing this task.
    \item I think my grammar/structure skills have improved after doing this task.
    \item I consulted the additional resources often.
    \item Additional resources are helpful for completing the task.
    \item I am willing to participate in this task again.
\end{itemize}

\subsection{Experiment Platform}
\begin{figure*}[ht!]
    \centering
    \begin{subfigure}{\textwidth}\centering
        \includegraphics[width=\textwidth]{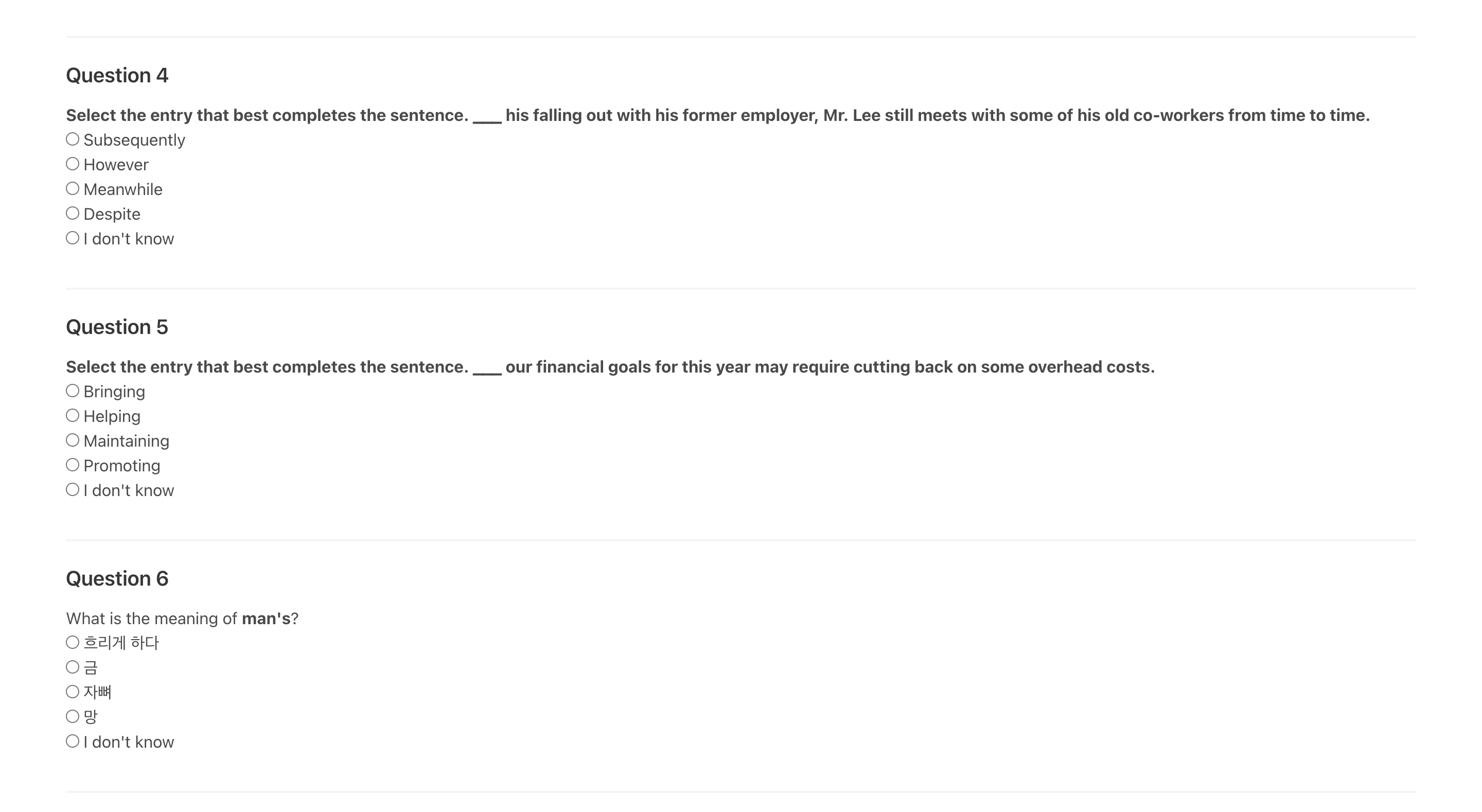}
        \caption{Pre-/post-test}
    \end{subfigure}
    \vspace{2mm}
    \begin{subtable}{\textwidth}\centering
        \includegraphics[width=\textwidth]{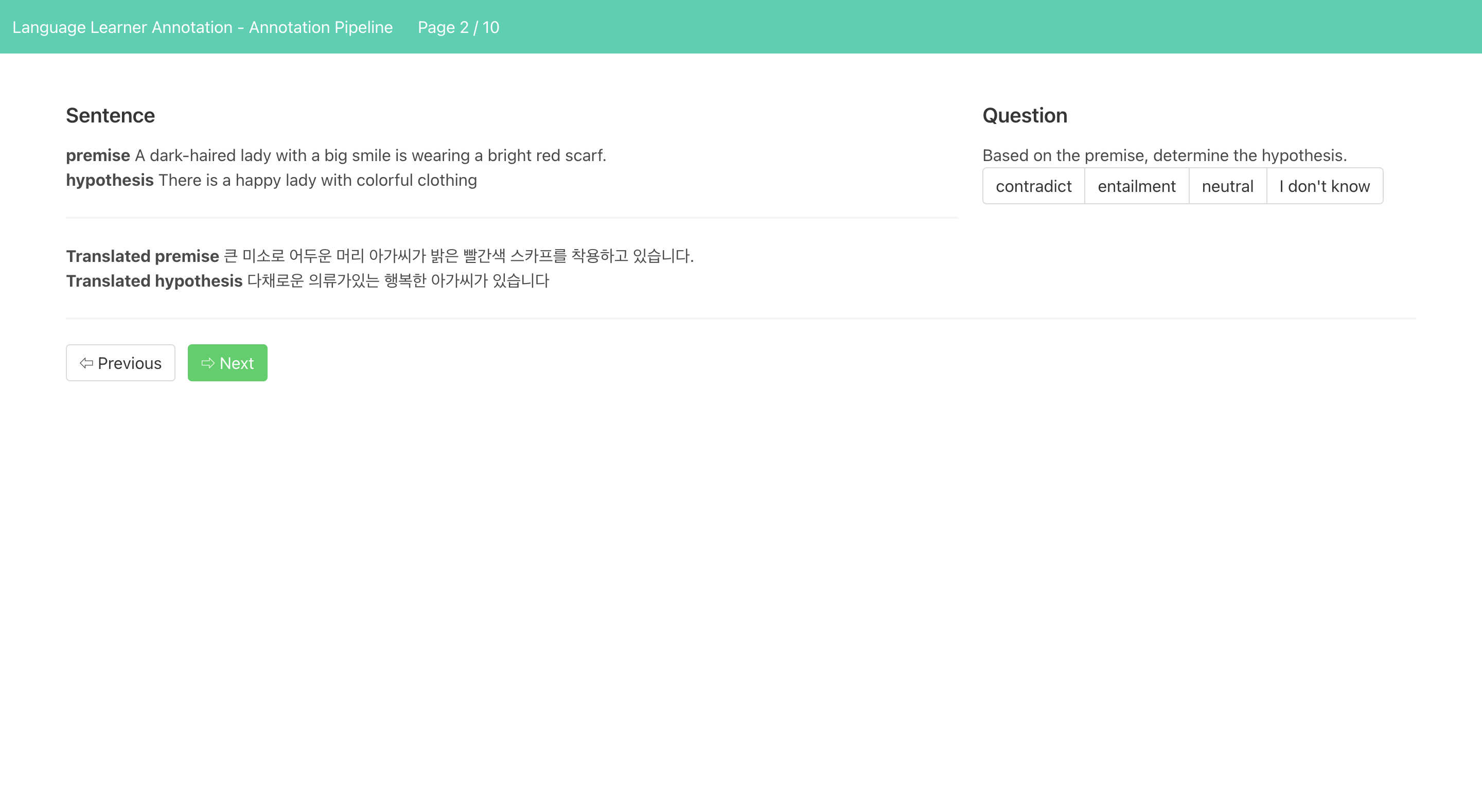}
    \caption{Annotation}
    \end{subtable}
    \caption{Screen shot of experiment platform}
    \label{fig:screenshot}
\end{figure*}

All experiments were done on the website that we made and all responses and time taken are recorded.
Figure \ref{fig:screenshot} shows the screenshots of the pre-/post-test (a) and annotation (b) steps.

\section{Further Results}
\label{appendix:annotation_accuracy}

\subsection{Annotation Quality}

\paragraph{Languages and Tasks.}
\begin{table}[h!]
    \begin{subtable}{\linewidth}\centering
        \begin{tabular}{@{}lccc@{}}
        \toprule
        \textbf{} & \textbf{EN}       & \textbf{KO}       & \textbf{ID}       \\ \midrule
        SA        & 7.87$_{\pm 0.30}$ & 7.59$_{\pm 0.21}$ & 8.35$_{\pm 0.16}$ \\
        NLI       & 6.81$_{\pm 0.20}$ & 6.54$_{\pm 0.18}$ & 6.46$_{\pm 0.26}$ \\
        NER       & 8.59$_{\pm 0.10}$ & 8.44$_{\pm 0.12}$ & 7.12$_{\pm 0.31}$ \\
        MRC       & 7.11$_{\pm 0.28}$ & 7.18$_{\pm 0.19}$ & 7.95$_{\pm 0.11}$ \\ \bottomrule
        \end{tabular}
        \caption{Annotation accuracy}
    \end{subtable}%
    
    \vspace{2mm}
    
    \begin{subtable}{\linewidth}\centering
        \begin{tabular}{@{}lccc@{}}
        \toprule
            & \textbf{EN}       & \textbf{KO}       & \textbf{ID}       \\ \midrule
        SA  & 0.60$_{\pm 0.03}$ & 0.62$_{\pm 0.02}$ & 0.71$_{\pm 0.03}$ \\
        NLI & 0.84$_{\pm 0.01}$ & 0.85$_{\pm 0.01}$ & 0.56$_{\pm 0.01}$ \\
        NER & 0.76$_{\pm 0.03}$ & 0.76$_{\pm 0.03}$ & 0.85$_{\pm 0.03}$ \\
        MRC & 0.63$_{\pm 0.03}$ & 0.51$_{\pm 0.03}$ & 0.53$_{\pm 0.04}$ \\ \bottomrule
        \end{tabular}
        \caption{Inter-annotator agreement measured by pairwise F1}
    \end{subtable}%
    
    \vspace{2mm}
    
    \begin{subtable}{\linewidth}\centering
        \begin{tabular}{@{}lccc@{}}
        \toprule
            & \textbf{EN}       & \textbf{KO}       & \textbf{ID}       \\ \midrule
        SA  & 3.34$_{\pm 0.34}$ & 3.28$_{\pm 0.25}$ & 2.43$_{\pm 0.18}$ \\
        NLI & 4.48$_{\pm 0.74}$ & 9.72$_{\pm 2.03}$ & 8.50$_{\pm 1.33}$ \\
        NER & 4.50$_{\pm 0.38}$ & 8.61$_{\pm 0.88}$ & 8.82$_{\pm 1.01}$ \\
        MRC & 6.29$_{\pm 0.57}$ & 9.58$_{\pm 0.72}$ & 7.27$_{\pm 0.41}$ \\ \bottomrule
        \end{tabular}
        \caption{Time spent (minutes)}
    \end{subtable}%
    
    \vspace{2mm}
    
    \begin{subtable}{\linewidth}\centering
        {\begin{tabular}{@{}lccc@{}}
        \toprule
            & \textbf{EN}       & \textbf{KO}       & \textbf{ID}       \\ \midrule
        SA  & 2.33$_{\pm 0.88}$ & 2.83$_{\pm 0.31}$ & 2.33$_{\pm 0.42}$ \\
        NLI & 2.60$_{\pm 0.60}$ & 3.17$_{\pm 0.48}$ & 4.00$_{\pm 1.00}$ \\
        NER & 3.40$_{\pm 0.60}$ & 3.17$_{\pm 0.48}$ & 3.50$_{\pm 0.50}$ \\
        MRC & 1.67$_{\pm 0.33}$ & 3.83$_{\pm 0.31}$ & 3.00$_{\pm 0.52}$ \\ \bottomrule
        \end{tabular}}
        \caption{Perceived difficulty from 1 \textit{(very easy)} to 5 \textit{(very hard)}}
    \end{subtable}
    
    \caption{Difficulty according to language and task}
    \label{tab:task_difficulty}
\end{table}

Table \ref{tab:task_difficulty} shows task difficulty with respect to four aspects: annotation accuracy, inter-annotator agreement, time, and perceived difficulty.

\paragraph{Majority Voted Labels.}
\begin{table*}[ht!]
\centering
\begin{tabular}{@{}llcccccc@{}}
\toprule
                                  &             & \multicolumn{3}{c}{\textbf{SA}}   & \multicolumn{3}{c}{\textbf{NLI}} \\ \cmidrule(l){3-8} 
                                  &             & EN         & KO        & ID       & EN       & KO        & ID        \\ \midrule
Native Speaker                    & -           & 0 / 12     & 0 / 12    & 0 / 7    & 7 / 35   & 11 / 38   & 1 / 10    \\ \midrule
\multirow{3}{*}{Language Learner} & All         & 0 / 16     & 0 / 20    & 0 / 13   & 0 / 31   & 0 / 21    & 0 / 28    \\
                                  & Dictionary  & 0 / 23     & 0 / 16    & 0 / 13   & 4 / 39   & 2 / 27    & 0 / 31    \\
                                  & Translation & 0 / 14     & 0 / 26    & 0 / 14   & 3 / 29   & 11 / 39   & 11 / 45   \\ \midrule
                                  &             & \multicolumn{3}{c}{\textbf{NER}}  & \multicolumn{3}{c}{\textbf{MRC}} \\ \cmidrule(l){3-8} 
                                  &             & EN         & KO        & ID       & EN       & KO        & ID        \\ \midrule
Native Speaker                    & -           & 3 / 21     & 1 / 19    & 2 / 18   & 4 / 23   & 6 / 19    & 3 / 20    \\ \midrule
\multirow{3}{*}{Language Learner} & All         & 4 / 17     & 6 / 21    & 3 / 20   & 5 / 22   & 7 / 24    & 6 / 25    \\
                                  & Dictionary  & 2 / 21     & 3 / 19    & 4 / 14   & 3 / 18   & 6 / 20    & 4 / 25    \\
                                  & Translation & 5 / 19     & 3 / 15    & 2 / 17   & 6 / 27   & 4 / 14    & 5 / 16    \\ \bottomrule
\end{tabular}
\caption{Majority vote results. (the number of splits / the number of incorrect) samples out of 120.}
\label{tab:aggregation_majority_vote}
\end{table*}

Table \ref{tab:aggregation_majority_vote} shows the statistics of aggregated labels using majority vote.
The number of splits means how many samples are not able to be aggregated in a single label (e.g., all annotators picked different labels, some annotators answered as \textit{I don't know} so that it was too few to be aggregated), and the number of incorrect samples means how many samples are different to the ground truth labels.

\paragraph{Additional Resources.}

\begin{table}[h!]
\centering
\begin{tabular}{@{}lcc@{}}
\toprule
\textbf{} & \textbf{Dictionary} & \textbf{Translation}  \\ \midrule
EN        & 7.21$_{\pm 0.24}$   & 7.84$_{\pm 0.12}$     \\
KO        & 7.77$_{\pm 0.12}$   & 7.11$_{\pm 0.15}$     \\
ID        & 8.14$_{\pm 0.10}$   & 7.05$_{\pm 0.17}$     \\ \bottomrule
\end{tabular}
\caption{Annotation accuracy with respect to languages and additional resources}
\label{tab:languages_vs_additional_resources}
\end{table}

Table \ref{tab:languages_vs_additional_resources} shows whether the types of additional resources that learners consult affect annotation accuracy among three languages.
English learners with the translation setting showed slightly better performance than those with the dictionary setting, while vice versa in Korean and Indonesian.
It implies that it would be better to provide translations in high-resource languages with reliable machine translation systems, while mid- to low-resource language learners should consult dictionaries.

\paragraph{Language Proficiency and Native Speakers.}

We recruited three native speakers of each language and asked them to do the same experiments (pre-test, annotation, and post-test).

\begin{table*}[ht!]
    \centering
    \begin{subtable}{\linewidth}\centering
        \begin{tabular}{@{}lcccc@{}}
        \toprule
            & \multicolumn{3}{c}{\textbf{Language Learners}}            & \textbf{Native Speakers}   \\ \cmidrule(l){2-4} \cmidrule(l){5-5}
            & Basic             & Intermediate      & Advanced          &                            \\ \midrule
        SA  & 6.96$_{\pm 0.25}$ & 8.26$_{\pm 0.16}$ & 8.43$_{\pm 0.16}$ & 9.00$_{\pm 0.14}$ \\
        NLI & 6.96$_{\pm 0.18}$ & 6.38$_{\pm 0.21}$ & 6.64$_{\pm 0.21}$ & 7.59$_{\pm 0.30}$ \\
        NER & 7.98$_{\pm 0.12}$ & 8.48$_{\pm 0.09}$ & 7.75$_{\pm 0.28}$ & 8.99$_{\pm 0.07}$ \\
        MRC & 6.56$_{\pm 0.26}$ & 7.88$_{\pm 0.12}$ & 7.71$_{\pm 0.13}$ & 8.51$_{\pm 0.14}$ \\ \bottomrule
        \end{tabular}
        \caption{Annotation accuracy}
    \end{subtable}%
    
    \vspace{2mm}
    
    \begin{subtable}{\linewidth}\centering
        \begin{tabular}{@{}lcccc@{}}
        \toprule
            & \multicolumn{3}{c}{\textbf{Language Learners}}             & \textbf{Native Speakers}   \\ \cmidrule(l){2-4} \cmidrule(l){5-5}
            & Basic             & Intermediate      & Advanced           &                            \\ \midrule
        SA  & 2.84$_{\pm 0.24}$ & 3.74$_{\pm 0.26}$ & 2.36$_{\pm 0.18}$  & 0.92$_{\pm 0.07}$        \\
        NLI & 8.58$_{\pm 0.69}$ & 5.78$_{\pm 1.23}$ & 10.92$_{\pm 3.46}$ & 3.38$_{\pm 0.59}$        \\
        NER & 7.95$_{\pm 0.51}$ & 5.83$_{\pm 0.50}$ & 7.68$_{\pm 1.12}$  & 7.23$_{\pm 2.90}$        \\
        MRC & 7.03$_{\pm 0.51}$ & 9.53$_{\pm 0.67}$ & 7.13$_{\pm 0.58}$  & 4.85$_{\pm 1.10}$        \\ \bottomrule
        \end{tabular}
        \caption{Time duration}
    \end{subtable}
    
    \caption{Task difficulty between three levels of learners and native speakers with respect to annotation accuracy and time duration}
    \label{tab:task_difficulty_native}
\end{table*}

Table \ref{tab:task_difficulty_native} shows the number of correct questions out of 10 and the time duration by each level of language learners and native speakers.
Native speakers achieved the highest accuracy across all tasks taking the shortest time.
It implies that there are some questions that native speakers can solve but learners cannot.
We discuss those samples in Section \ref{sample-level_analysis}.
Time duration shows a significant gap between learners and native speakers, especially on NLI, but the gap was minimized at NER whose task requires annotators to tag all sequences.

\subsection{Training Simulation with Learners' Annotation}

\paragraph{Soft-labeled Synthetic Data.}

\begin{table*}[!ht]
\centering
\resizebox{\textwidth}{!}{
\begin{tabular}{@{}llcccccc@{}}
\toprule
                                   &             & \multicolumn{3}{c}{\textbf{SA}}                                          & \multicolumn{3}{c}{\textbf{NLI}}                                      \\ \cmidrule(l){3-5} \cmidrule(l){6-8}
                                   &             & EN                   & KO                        & ID                    & EN                    & KO                    & ID                    \\ \midrule
Ground Truth                       & -           & 89.56$_{\pm 1.11}$   & 85.29$_{\pm 0.79}$        & 97.20$_{\pm 0.86}$    & 79.05$_{\pm 1.44}$    & 79.00$_{\pm 2.48}$    & 68.20$_{\pm 1.32}$  \\
MT Dataset                         & -           & 79.25$_{\pm 1.25}$   & 75.27$_{\pm 1.33}$        & 87.19$_{\pm 1.39}$    & 56.78$_{\pm 2.26}$    & 47.06$_{\pm 1.26}$    & 52.35$_{\pm 1.35}$                  \\ \midrule
Native Speakers                    & -           & 70.66$_{\pm 1.60}$   & 84.66$_{\pm 1.40}$        & 96.48$_{\pm 0.76}$    & 67.67$_{\pm 1.60}$    & 56.18$_{\pm 1.55}$    & 67.12$_{\pm 1.61}$  \\ \midrule
\multirow{3}{*}{Language Learners} & All         & 85.75$_{\pm 2.21}$   & 80.22$_{\pm 0.92}$        & 92.37$_{\pm 1.45}$    & 78.38$_{\pm 2.05}$    & 72.51$_{\pm 1.22}$    & 61.99$_{\pm 3.18}$  \\
                                   & Dictionary  & 77.35$_{\pm 1.92}$   & 82.94$_{\pm 1.09}$        & 91.04$_{\pm 0.65}$    & 62.40$_{\pm 2.86}$    & 70.27$_{\pm 1.89}$    & 63.33$_{\pm 2.24}$    \\
                                   & Translation & 85.29$_{\pm 1.28}$   & 72.98$_{\pm 1.74}$        & 90.40$_{\pm 1.07}$    & 68.88$_{\pm 1.68}$    & 65.61$_{\pm 1.06}$    & 56.54$_{\pm 3.86}$    \\ \bottomrule

\end{tabular}
}
\caption{Experimental results of BERT-based models trained on labels generated or synthesized by each group using soft-labeling}
\label{tab:synthetic_result_soft_label}
\end{table*}

We tried training simulations with BERT-based models on synthetic data generated using soft labeling.
We used soft labeling instead of majority voting to consider the variance among the annotators.
Table \ref{tab:synthetic_result_soft_label} shows experimental results of BERT-based models on synthetic data whose data distributions come from the soft-labeled aggregations.
It delivers similar findings to Table \ref{tab:synthetic_result_majority_vote}, while showing some noises.
Models trained on native speakers' synthetic labels sometimes achieved similar performance to the Ground Truth while sometimes achieving the poorest performance such as EN-SA, EN-NLI, and KO-NLI.
Our native annotators showed low inter-annotator agreement in those languages and tasks, so the synthetic labels based on native speakers' annotations were noisy.

\paragraph{Few-shot Learning using mT5.}

\begin{table*}[!ht]
\centering
\resizebox{\textwidth}{!}{
\begin{tabular}{@{}llcccccc@{}}
\toprule
                                   &             & \multicolumn{3}{c}{\textbf{SA}}                              & \multicolumn{3}{c}{\textbf{NLI}}                             \\ \cmidrule(l){3-5} \cmidrule(l){6-8}
                                   &             & EN                 & KO                 & ID                 & EN                 & KO                 & ID                 \\ \midrule
Ground Truth                       & -           & 89.07$_{\pm 3.45}$ & 89.19$_{\pm 2.85}$ & 94.07$_{\pm 3.62}$ & 78.34$_{\pm 2.47}$ & 80.64$_{\pm 3.20}$ & 68.34$_{\pm 2.82}$ \\
MT Dataset                         & -           & 85.13$_{\pm 2.12}$ & 84.57$_{\pm 3.18}$ & 90.10$_{\pm 2.53}$ & 74.48$_{\pm 3.79}$ & 77.92$_{\pm 2.29}$ & 63.20$_{\pm 2.95}$ \\ \midrule
Native Speakers                    & -           & 88.64$_{\pm 3.11}$ & 88.67$_{\pm 3.54}$ & 93.64$_{\pm 2.12}$ & 77.45$_{\pm 2.85}$ & 79.36$_{\pm 3.13}$ & 66.45$_{\pm 3.58}$ \\ \midrule
\multirow{3}{*}{Language Learners} & All         & 87.26$_{\pm 3.10}$ & 87.32$_{\pm 2.56}$ & 93.26$_{\pm 3.13}$ & 78.41$_{\pm 3.46}$ & 80.82$_{\pm 2.70}$ & 68.41$_{\pm 3.15}$ \\
                                   & Dictionary  & 88.16$_{\pm 3.55}$ & 88.28$_{\pm 3.53}$ & 94.16$_{\pm 2.13}$ & 76.64$_{\pm 3.39}$ & 81.08$_{\pm 3.75}$ & 69.64$_{\pm 2.76}$ \\
                                   & Translation & 85.39$_{\pm 2.71}$ & 87.47$_{\pm 2.34}$ & 92.47$_{\pm 2.88}$ & 74.69$_{\pm 2.99}$ & 73.03$_{\pm 2.65}$ & 69.84$_{\pm 3.19}$ \\ \bottomrule
\end{tabular}
}
\caption{Experimental results of Few-shot Learning using mT5}
\label{tab:few_shot_result_mt5}
\end{table*}

We also tried few-shot learning with mT5\textsubscript{BASE} \citep{xue-etal-2021-mt5}, a large-scale multilingual pretrained model which covers 101 languages including our target languages: English, Korean, and Indonesian.
Table \ref{tab:few_shot_result_mt5} shows that all models achieved comparable results to the baseline model within the margin of error.
The gap among all models was relieved and we suppose that large-scale LMs with massive training data, including mT5, can perform too well on our common NLP tasks and our labeled data were too small to affect those models.

\section{Further Discussions}

\subsection{Learning Effect}

\paragraph{Additional Resources.}

\begin{table}[h!]
\centering
\begin{subtable}{\columnwidth}\centering
    \begin{tabular}{@{}lcc@{}}
    \toprule
    \textbf{} & \textbf{Dictionary} & \textbf{Translation} \\ \midrule
    pre-test  & 3.63$_{\pm 0.07}$   & 3.32$_{\pm 0.07}$    \\
    post-test & 3.53$_{\pm 0.07}$   & 3.41$_{\pm 0.06}$    \\ \bottomrule
    \end{tabular}
    \caption{Number of correct standardized test questions out of 5}
\end{subtable}%

\vspace{2mm}

\begin{subtable}{\columnwidth}\centering
    \begin{tabular}{@{}lcc@{}}
    \toprule
    \textbf{} & \textbf{Dictionary} & \textbf{Translation} \\ \midrule
    pre-test  & 8.81$_{\pm 0.08}$   & 8.47$_{\pm 0.08}$    \\
    post-test & 9.29$_{\pm 0.07}$   & 8.92$_{\pm 0.07}$    \\ \bottomrule
    \end{tabular}
    \caption{Number of correct word meaning questions out of 10}
\end{subtable}

\caption{Effect of additional resources in language learning with respect to language proficiency}
\label{tab:additional_resources_pre_post_test}
\end{table}

Table \ref{tab:additional_resources_language_proficiency} (b) shows that both additional resources helped learners to remind or learn vocabulary used in the annotation samples.

\paragraph{Perceived Learning Effect.}

\begin{table}[h!]
\centering
\resizebox{\columnwidth}{!}{
\begin{tabular}{@{}lccc@{}}
\toprule
                        & \textbf{Basic}    & \textbf{Intermediate} & \textbf{Advanced} \\ \midrule
vocab   & 4.21$_{\pm 0.13}$ & 3.41$_{\pm 0.13}$     & 3.40$_{\pm 0.13}$ \\
grammar & 3.36$_{\pm 0.13}$ & 2.77$_{\pm 0.13}$     & 2.65$_{\pm 0.13}$ \\
willingness             & 3.93$_{\pm 0.21}$ & 2.95$_{\pm 0.21}$     & 3.30$_{\pm 0.21}$ \\ \bottomrule
\end{tabular}
}
\caption{Users' responses on post-survey in terms of learning effect on vocabulary and grammar and willingness to re-participate}
\label{tab:learning_effect_willingness}
\end{table}

Table \ref{tab:learning_effect_willingness} shows similar trends to the previous results that basic-level learners perceived more learning effects on both vocabulary and grammar.
They tend to show more willingness to re-participate in data annotation.
Advanced-level learners show a high willingness to re-participate in data annotation, and this is because it was hard to improve their language proficiency. However, the sentences in data annotation were easy enough for them.

\begin{table}[h!]
\centering
\resizebox{\columnwidth}{!}{
\begin{tabular}{@{}lccc@{}}
\toprule
            & \textbf{Basic}    & \textbf{Intermediate} & \textbf{Advanced} \\ \midrule
pre-survey  & 0.57$_{\pm 0.14}$ & 2.45$_{\pm 0.17}$     & 3.20$_{\pm 0.14}$ \\
post-survey & 1.29$_{\pm 0.19}$ & 2.55$_{\pm 0.14}$     & 3.20$_{\pm 0.17}$ \\ \bottomrule
\end{tabular}
}
\caption{Self-rated language proficiency before and after data annotation experiment}
\label{tab:diff_on_self-rated_language_fluency}
\end{table}

Table \ref{tab:diff_on_self-rated_language_fluency} shows self-rated language proficiency before and after the experiments when the description of CEFR criteria was given.
Basic-level learners felt that their language proficiency had improved, while other levels of learners did not show a significant difference.
Advanced-level learners tend to underestimate their language proficiency humbly.

\paragraph{Language Proficiency and Additional Resources.}

\begin{table}[h!]
\centering
\begin{subtable}{\columnwidth}\centering
    \begin{tabular}{@{}lcc@{}}
    \toprule
    \textbf{}       & \textbf{Dictionary} & \textbf{Translation}  \\ \midrule
    Basic           & 7.40$_{\pm 0.18}$   & 6.90$_{\pm 0.17}$     \\
    Intermediate    & 7.86$_{\pm 0.12}$   & 7.60$_{\pm 0.13}$     \\
    Advanced        & 7.99$_{\pm 0.14}$   & 7.31$_{\pm 0.16}$     \\ \bottomrule
    \end{tabular}
    \caption{Annotation accuracy}
\end{subtable}%

\vspace{2mm}

\begin{subtable}{\columnwidth}\centering
    \begin{tabular}{@{}lcc@{}}
    \toprule
                    & \textbf{Dictionary} & \textbf{Translation}  \\ \midrule
    Basic           & 4.67$_{\pm 0.21}$   & 3.25$_{\pm 0.59}$     \\
    Intermediate    & 2.75$_{\pm 0.45}$   & 2.70$_{\pm 0.42}$     \\
    Advanced        & 2.75$_{\pm 0.41}$   & 2.75$_{\pm 0.39}$     \\ \bottomrule
    \end{tabular}
    \caption{Frequency of consulting additional resources}
\end{subtable}

\vspace{2mm}

\begin{subtable}{\columnwidth}\centering
    \begin{tabular}{@{}lcc@{}}
    \toprule
                    & \textbf{Dictionary} & \textbf{Translation}  \\ \midrule
    Basic           & 4.67$_{\pm 0.21}$   & 3.75$_{\pm 0.53}$     \\
    Intermediate    & 3.42$_{\pm 0.47}$   & 3.30$_{\pm 0.37}$     \\
    Advanced        & 3.62$_{\pm 0.46}$   & 3.25$_{\pm 0.41}$     \\ \bottomrule
    \end{tabular}
    \caption{Help of additional resources}
\end{subtable}

\caption{Effect of additional resources with respect to language proficiency}
\label{tab:additional_resources_language_proficiency}
\end{table}

Table \ref{tab:additional_resources_language_proficiency} (a) shows annotation accuracy compared to the ground truth labels concerning the learners' language proficiency level and the additional resources they used.
There was no significant difference between the two settings with the learners either in the intermediate or the advanced level, while basic level learners achieved higher accuracy in dictionary settings.
We suppose that basic-level learners might not be able to fill the gap of the wrong spans in the machine-translated sentence.

Table \ref{tab:additional_resources_language_proficiency} (b)-(c) show users' responses on how frequently they consult additional resources and how helpful they were in data annotation.
The frequency that the learners consult the additional resources and how the additional resources are helpful go together.
All levels of learners replied that the dictionary setting was more helpful than the translation setting.
Most basic-level learners in all languages consult and rely on additional resources.

There was no significant trend in the learners' frequency of consulting the additional resources concerning language and types of additional resources.
Still, learners of all languages replied that the dictionary setting was more helpful for data annotation than the translation setting.

\subsection{Feedback from Participants}

Table \ref{tab:task_difficulty} (c) shows perceived difficulty based on users' responses on post-survey.
Participants responded that NER was the most complicated task and SA was the easiest.
This result looks awkward considering that language learners achieved the highest accuracy in NER.

Learners replied that exactly distinguishing the start and the end of the named entity was confused in NER, and some named entities were unfamiliar with them if they were not used to the domain.
All learners in the translation-provided setting on NLI replied that the machine-translated sentences were incorrect and even disturbing to infer the textual entailment between two sentences.
Most Indonesian learners on SA replied that the sentences usually contain multiple sentiments, representing that some points are good, but others are bad, so they are unsure about their labels.
This is probably due to the characteristics of IndoLEM \citep{koto-etal-2020-indolem} whose sentences come from Hotel reviews with multiple features.
Learners should read a passage in MRC so that it helps to improve their language proficiency, while advanced-level learners who are fluent in the target language replied that they do not have to read the whole passage but read the sentence that contains the answer span.

\section{Qualitative Analysis}

\subsection{Failure Reason Analysis on Learners' Annotation}

Table \ref{tab:sample_level_analysis_example} shows the examples of three failure reasons: ungrammatical sentence, task ambiguity, and culturally-nuanced expression.
Missing period between two short sentences in the SA sample (a) leads to misunderstandings among learners.
Ambiguity, whether ``\textit{people}'' and ``\textit{some people}'' in premise and hypothesis are indicating the same in (b), leads all learners and native speakers to get confused between \texttt{neutral} and \texttt{contradiction}, which is an ambiguity of NLI itself.
``\textit{ajaran yang dipercayai}'' in questions in the MRC sample (c) literally means ``\textit{teachings believed by}'' in Indonesian, but its correct translation is ``\textit{belief}'' or ``\textit{religion}''.
Learners failed to interpret those difficult and culturally-nuanced expressions correctly and generated wrong labels, while all native speakers found the same answer.

\subsection{Qualitative Analysis on Pre-/Post-test}

We analyze the characteristics of pre- and post-test questions that the learners get wrong. For English, two questions that every learner got wrong were GRE questions, which are notably difficult even for native speakers. Many learners picked the ``\textit{I don't know}'' option for GRE questions as well. For Korean, there was no question that every learner got wrong. However, for A-level learners, a large number of them answered `Arrange the sentences in the correct order' questions incorrectly. 
The difficulty may stem from their insufficient knowledge of transition signals and the logical flow in the target language. 
Also, learners chose ``\textit{I don't know}'' option a lot for questions requiring an understanding of newspaper titles. For Indonesian, learners mostly fail on questions related to prepositions, prefixes and suffixes, and formal word formation. 

Most of the questions that most learners answered incorrectly require an understanding of the context and the grammatical structure. These aspects of language are difficult to learn within a short time, attributing to the insignificant difference in the scores between the pre- and post-tests.

\clearpage
\begin{sidewaystable*}
\small
\begin{tabularx}{\textheight}{@{}cccrXcccp{0.08\textheight}@{}}
\toprule
                     & Lang.                & Level              & Type                                                                     & Sentence                                                                                                                                                                                                                                                                                                                                                                                                                                                                 & Ground Truth                          & Language Learners                                                                         & Native Speakers                                                                          & Failure Reason                                                                                                                   \\ \midrule
\multirow{3}{*}{(a)} & \multirow{3}{*}{KO} & \multirow{3}{*}{1} & Original                                                                 & \begin{CJK}{UTF8}{mj}스토리가 어려움 볼만함\end{CJK}                                                                                                                                                                                                                                                                                                                                                                                                                               & \multirow{3}{*}{pos}        & \multirow{3}{*}{neg}                                                            & \multirow{3}{*}{pos}                                                            & Ungrammatical sentence \\ \cmidrule(lr){4-5}
                     &                     &                    & \begin{tabular}[c]{@{}r@{}}Machine\\ Trans\end{tabular}                  & Story is difficult to see                                                                                                                                                                                                                                                                                                                                                                                                                                                &                             &                                                                                 &                                                                                 &                                                                                                                                  \\ \cmidrule(lr){4-5}
                     &                    &                    & \begin{tabular}[c]{@{}r@{}}Correct\\ Trans\end{tabular}                  & The story if difficult, [but it's] worth watching.                                                                                                                                                                                                                                                                                                                                                                                                                         &                             &                                                                                 &                                                                                 &                                                                                                                                  \\ \midrule
\multirow{6}{*}{(b)} & \multirow{6}{*}{EN} & \multirow{6}{*}{2} & \multirow{2}{*}{Original}                                                & [Premise] People standing at street corner in France.                                                                                                                                                                                                                                                                                                                                                                                                                    & \multirow{6}{*}{con}        & \multirow{6}{*}{neu}                                                            & \multirow{6}{*}{neu}                                                            & \multirow{6}{*}{Task ambiguity}                                                                                                  \\
                     &                    &                    &                                                                          & [Hypothesis] Some people are taking a tour of the factory.                                                                                                                                                                                                                                                                                                                                                                                                               &                             &                                                                                 &                                                                                 &                                                                                                                                  \\ \cmidrule(lr){4-5}
                     &                    &                    & \multirow{2}{*}{\begin{tabular}[c]{@{}r@{}}Machine\\ Trans\end{tabular}} & \begin{CJK}{UTF8}{mj}[Premise] 프랑스의 거리 모퉁이에 서있는 사람들.\end{CJK}                                                                                                                                                                                                                                                                                                                                                                                                            &                             &                                                                                 &                                                                                 &                                                                                                                                  \\
                     &                    &                    &                                                                          & \begin{CJK}{UTF8}{mj}[Hypothesis] 어떤 사람들은 공장을 여행하고 있습니다.\end{CJK}                                                                                                                                                                                                                                                                                                                                                                                                        &                             &                                                                                 &                                                                                 &                                                                                                                                  \\ \cmidrule(lr){4-5}
                     &                    &                    & \begin{tabular}[c]{@{}r@{}}Correct\\ Trans\end{tabular}                  & \begin{CJK}{UTF8}{mj}[Premise] 프랑스의 거리 모퉁이에 서있는 사람들.\end{CJK}                                                                                                                                                                                                                                                                                                                                                                                                            &                             &                                                                                 &                                                                                 &                                                                                                                                  \\
                     &                    &                    &                                                                          & \begin{CJK}{UTF8}{mj}[Hypothesis] 어떤 사람들은 공장을 관광하고 있습니다.\end{CJK}                                                                                                                                                                                                                                                                                                                                                                                                        &                             &                                                                                 &                                                                                 &                                                                                                                                  \\ \midrule
\multirow{6}{*}{(c)} & \multirow{6}{*}{ID} & \multirow{6}{*}{5} & \multirow{2}{*}{Original}                                                & [Context] ... Mereka membangun komunitas dengan berpegang teguh pada spiritualitas sebagai dasar pembentukan ajarannya. Tidak jarang pula mereka menyebut kepercayaannya sebagai agama Jawa. Melalui kepercayaan ini, mereka melakukan penggalian kembali kepercayaan dan nilai-nilai spiritualitas masyarakat Jawa masa lalu, terutama pada masa prapatrimonial. ... & \multirow{6}{*}{agama Jawa} & \multirow{6}{*}{\begin{tabular}[c]{@{}@{}c@{}}\textit{I don't know}; \\ spiritualitas; \\ etc \\\end{tabular}} & \multirow{6}{*}{agama Jawa} & \multirow{6}{*}{\begin{tabular}[c]{p{0.08\textwidth}}Culturally-nuanced expression\end{tabular}}                                                                                                \\
                     &                    &                    &                                                                          & [Question] Apakah ajaran yang dipercayai Suku Dayak Hindu Budha Bumi Segandu Indramayu?                                                                                                                                                                                                                                                                                                                                                                                              &                             &                                                                                 &                                                                                 &                                                                                                                                  \\ \cmidrule(lr){4-5}
                     &                    &                    & \multirow{2}{*}{\begin{tabular}[c]{@{}r@{}}Machine\\ Trans\end{tabular}} & [Context] ... They built the community by clinging to spirituality as the basis for the formation of his teachings. Not infrequently also they mentioned his belief as Java religion.Through this belief, they re-excavated trust and the spirituality values of the past Javanese society, especially during the praprimonial period. ...                                            &                             &                                                                                 &                                                                                 &                                                                                                                                  \\
                     &                    &                    &                                                                          & [Question] What is the teachings believed by the Hindu Buddhist Bumi Division of Indramayu?                                                                                                                                                                                                                                                                                                                                                                                                      &                             &                                                                                 &                                                                                 &                                                                                                                                  \\ \cmidrule(lr){4-5}
                     &                    &                    & \multirow{2}{*}{\begin{tabular}[c]{@{}r@{}}Correct\\ Trans\end{tabular}} & [Context] ... They built the community by clinging to spirituality as the foundation for their teachings formation. Not infrequently, they called their belief as Java religion. Through this belief, they re-dig trust and the spiritual values of the past Javanese society, especially during the pre-patrimonial period. ...                               &                             &                                                                                 &                                                                                 &                                                                                                                                  \\
                     &                    &                    &                                                                          & [Question] What is the teachings believed by \textit{(belief/religion)} the Dayak Hindu Buddha Bumi Segandu Indramayu?                                                                                                                                                                                                                                                                                                                                                                                            &                             &                                                                                 &                                                                                 &                                                                                                                                  \\ \bottomrule
\end{tabularx}
\caption{Example annotation questions that all learners fail}
\label{tab:sample_level_analysis_example}
\end{sidewaystable*}

\end{document}